\documentclass[preprint,12pt]{article}
\usepackage[a4paper,margin=1in]{geometry}
\usepackage{cite}
\usepackage{amsmath,amssymb,amsfonts}
\usepackage{algorithmic}
\usepackage{graphicx}
\usepackage{algorithm,algorithmic}
\usepackage{hyperref}
\usepackage{textcomp}
\usepackage{authblk}
\usepackage{adjustbox}
\usepackage{multirow}
\usepackage{multicol}
\usepackage{booktabs}
\usepackage{xcolor}
\usepackage{colortbl}

\usepackage{mathtools}
\usepackage{comment}

\usepackage{subcaption}

\providecommand{\keywords}[1]{\textbf{\textit{Keywords---}} #1}

\begin{document}
\title{Bayesian Integration of Nonlinear Incomplete Clinical Data}

\author[1]{Lucía González-Zamorano$^{*,\dagger,}$}
\author[1]{Nuria Balbás-Esteban$^{\dagger,}$}
\author[1,2]{Vanessa Gómez-Verdejo}
\author[3]{Albert Belenguer-Llorens}
\author[1]{Carlos Sevilla-Salcedo}

\affil[1]{Department of Signal Theory and Communications, Universidad Carlos III de Madrid, Leganés, 28911, Spain}
\affil[2]{Instituto de Investigación Sanitaria Gregorio Marañón (IiSGM), Madrid, 28009, Spain}
\affil[3]{Department of Health Sciences and Technology (DHEST), ETH Zurich, 8092, Zürich, Switzerland}

\setcounter{footnote}{0}
\footnotetext[1]{Corresponding author (e-mail: \texttt{lgzamorano@tsc.uc3m.es})}
\footnotetext[2]{Equal contribution}


\date{
\footnotesize
}


\maketitle

\begin{abstract}
Multimodal clinical data are characterized by high dimensionality, heterogeneous representations, and structured missingness, posing significant challenges for predictive modeling, data integration, and interpretability. We propose BIONIC (Bayesian Integration of Nonlinear Incomplete Clinical data), a unified probabilistic framework that integrates heterogeneous multimodal data under missingness through a joint generative-discriminative latent architecture. BIONIC uses pretrained embeddings for complex modalities such as medical images and clinical text, while incorporating structured clinical variables directly within a Bayesian multimodal formulation. The proposed framework enables robust learning in partially observed and semi-supervised settings by explicitly modeling modality-level and variable-level missingness, as well as missing labels. We evaluate BIONIC on three multimodal clinical and biomedical datasets, demonstrating strong and consistent discriminative performance compared to representative multimodal baselines, particularly under incomplete data scenarios. Beyond predictive accuracy, BIONIC provides intrinsic interpretability through its latent structure, enabling population-level analysis of modality relevance and supporting clinically meaningful insight.

\end{abstract}


\section{Introduction}
\label{sec:introduction}

Clinical decision-making increasingly relies on the integration of heterogeneous data sources, including medical imaging, unstructured clinical text, and structured clinical records~\cite{ektefaie2023multimodal}. In real-world clinical settings, these modalities are acquired unevenly across patients, and are often available for limited cohort sizes, leading to highly heterogeneous, sparse, and small-sample multimodal datasets. In many cases, entire data views are missing for a given patient: imaging studies may not be performed due to contraindications or limited availability, and unstructured reports may be unavailable outside specialized care. As a result, patients are frequently characterized by partially observed multimodal configurations rather than complete modality sets.
Crucially, this view-level missingness is not incidental; it reflects resource constraints and cohort-specific inclusion criteria, and is therefore informative of both patient state and care pathways~\cite{austin2021missing,heymans2022handling}. Multimodal clinical datasets thus differ from many benchmark datasets, which typically assume large sample sizes and randomly missing views, and therefore fail to capture the low-sample, decision-driven missingness patterns that characterize real clinical workflows.


For these reasons, most multimodal learning approaches struggle in clinically realistic settings. Early fusion strategies~\cite{imrie2025automated} require complete modality availability, while late fusion methods ~\cite{imrie2025automated} often fail to capture meaningful cross-modal interactions when modalities are inconsistently observed. Transformer-based multimodal architectures typically assume synchronized and densely observed inputs~\cite{ma2022multimodal} and mechanisms for handling missing data are often limited to variable-level masking within individual modalities or to heuristic view imputation strategies, while principled modeling of view-level missingness remains an open and challenging problem in multimodal learning~\cite{lowquality2023survey}. Treating missingness as a preprocessing problem through ad-hoc imputation can therefore introduce bias~\cite{sterne2009multiple}, distort cross-modal relationships~\cite{karanth2024systematic}, and obscure clinically meaningful dependencies~\cite{austin2021missing}.

A further challenge arises from the mismatch between the high dimensionality of clinical data and the limited size of available cohorts. Medical images, histopathology slides, and molecular profiles are intrinsically high-dimensional; yet, clinical datasets often remain modest due to recruitment constraints and acquisition costs~\cite{cao2024small,button2013power}. Pretrained Foundation Models (FMs) have become increasingly prevalent in biomedical applications to mitigate this imbalance, providing compact and transferable representations learned from large external datasets. General-purpose models (e.g., CLIP~\cite{radford2021learning}, LLaVA~\cite{liu2023visual}) as well as domain-specific medical encoders (e.g., MedicalNet~\cite{chen2019med3d}, CLAM~\cite{lu2021clam}) for imaging and clinical text are now routinely used to extract fixed embeddings that serve as inputs to downstream predictive models.

However, when these embeddings are employed as black-box feature extractors within purely discriminative pipelines, critical limitations remain. First, such approaches typically treat each modality independently and do not explicitly model structured missingness at the view level, relying instead on ad-hoc imputation or modality-specific heuristics~\cite{wang2024deep}. Second, they offer limited support for principled modality or feature selection, making it difficult to assess which data sources drive predictions or to adapt models to partially observed multimodal configurations. Finally, uncertainty estimates (when provided at all) are often poorly calibrated~\cite{guo2017calibration,laves2020well} and post-hoc explanation methods (e.g., SHAP~\cite{lundberg2017unified}, LIME~\cite{ribeiro2016should}) applied on top of discriminative classifiers may lack global consistency and reliability~\cite{salih2024perspective}. Together, these limitations hinder the safe and trustworthy deployment of multimodal models in clinical settings, where interpretability and uncertainty quantification are as critical as predictive accuracy.

To address these limitations, we propose BIONIC (Bayesian Integration of Nonlinear Incomplete Clinical data), a unified probabilistic framework for multimodal clinical learning under structured missingness. Building on a Bayesian multi-view model previously validated for incomplete clinical data~\cite{belenguer2025interpretable}, BIONIC operates directly on heterogeneous fixed-dimensional representations produced by pretrained FMs. Rather than learning from raw high-dimensional inputs, the framework employs frozen encoders to obtain nonlinear embeddings offline, which are then integrated through a generative–discriminative architecture. Linear projections combined with sparsity-inducing priors allow BIONIC to automatically adapt the effective dimensionality of pretrained embeddings to the size and structure of the available data, yielding compact and task-relevant representations that improve robustness in limited-cohort settings. Missing variables, entire modalities, and labels are explicitly modeled within a shared latent space, enabling probabilistic imputation and semi-supervised learning without ad-hoc preprocessing.

This work makes four main contributions. First, we formulate multimodal clinical prediction as a Bayesian learning problem that integrates heterogeneous structured data with embeddings derived from unstructured modalities. Second, we introduce a unified treatment of missingness at the variable, view, and label levels, enabling principled imputation and semi-supervised learning consistent with real-world clinical acquisition processes. Third, we propose an intrinsic and analytically tractable interpretability mechanism derived from the linear discriminative pathway, enabling propagation of relevance from the latent space to embedding representations and, when supported by pretrained decoders, back to the input domains. Finally, we adopt a fully Bayesian formulation with sparsity-inducing priors that automatically controls representation complexity, improving robustness and generalization when learning from high-dimensional pretrained embeddings.

We evaluate BIONIC on real-world multimodal oncology datasets through three complementary experimental studies. First, we compare its discriminative performance against state-of-the-art multimodal baselines. Second, we analyze the benefits of the generative formulation under three supervision regimes, highlighting its robustness to missing modalities and labels. Finally, we conduct an interpretability analysis demonstrating how the proposed framework yields a principled sensitivity metric that links embedding- or feature-level representations to the downstream prediction task.

We consider a multimodal clinical classification problem with $N$ samples observed across $M$ heterogeneous modalities. The dataset is represented as $\{(\{\mathbf{x}_{n}^{(m)}\}_{m=1}^M, \mathbf{t}_{n})\}_{n=1}^N$, where $\mathbf{x}_{n}^{(m)} \in \mathbb{R}^{D^{(m)}}$ denotes the representation of modality $m \in \mathcal{M}=\{1,\dots,M\}$ and $\mathbf{t}_{n} \in \{0,1\}^{C}$ is its one-hot encoded label over $C$ classes when available. Let $\mathcal{M}_a \subseteq \mathcal{M}$ denote the set of views that are available (fully or partially) for patient $n$.




BIONIC (Bayesian Integration of Nonlinear Incomplete Clinical data) is a probabilistic framework for multimodal learning with missing data. Building on a Bayesian multi-view model previously validated for incomplete data~\cite{belenguer2025interpretable}, it operates directly on heterogeneous fixed-dimensional representations produced by pretrained FMs, enabling multimodal integration without retraining complex encoders. The framework employs a dual latent structure that separates generative and discriminative roles. The generative latent space captures shared and modality-specific variability and provides a principled mechanism for handling missing variables or entire modalities, while the discriminative latent space aggregates task-relevant information for prediction. 

\subsection{Probabilistic modeling and inference}
\label{sec:proba}
\begin{figure*}
    \centering
    \includegraphics[width=0.7\linewidth]{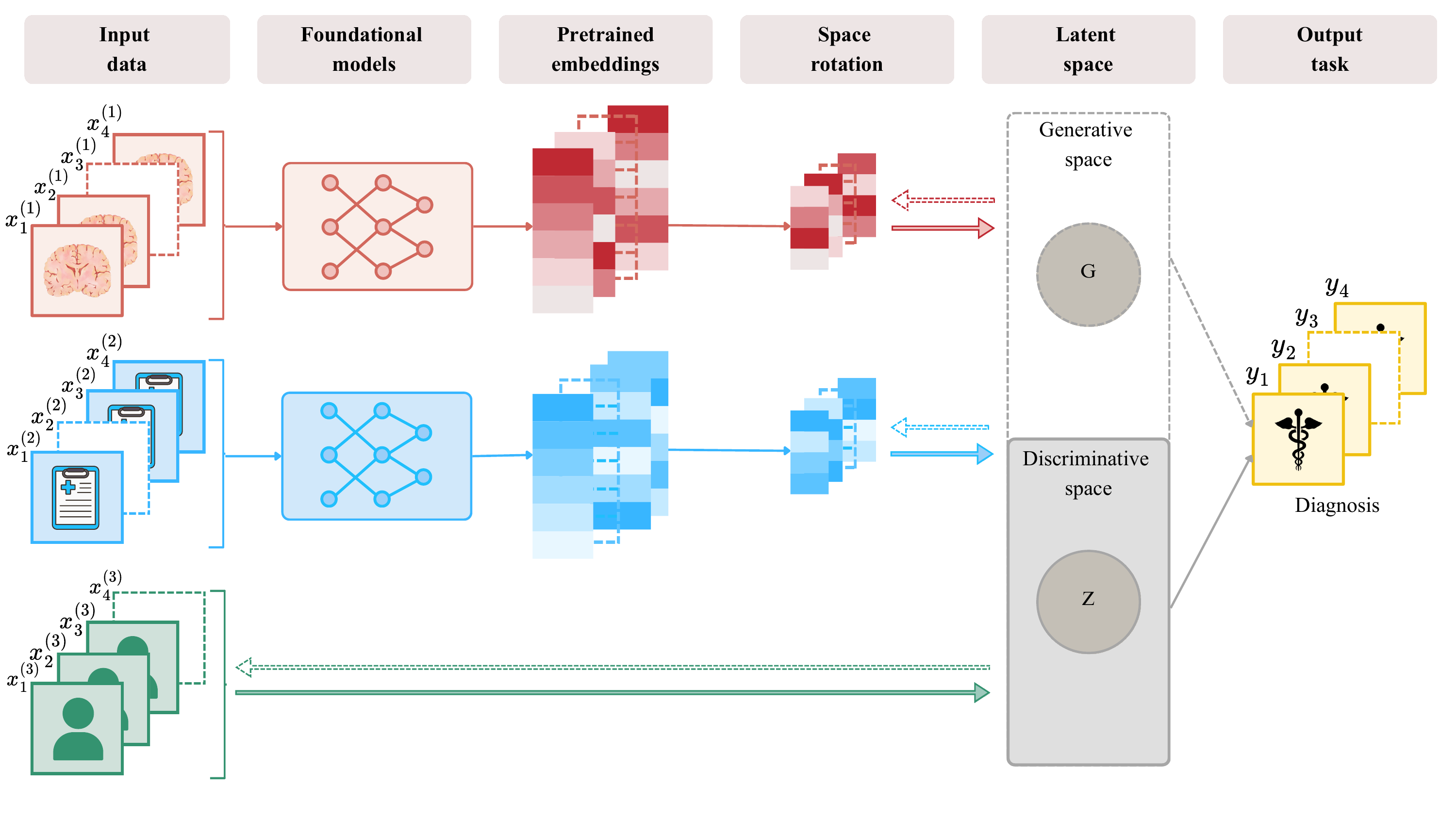}
    \caption{Overview of the proposed BIONIC framework. Structured clinical variables and pretrained embeddings are mapped through view-specific projections into a dual latent space, where a generative component models multimodal structure and missing data, and a discriminative component supports the downstream prediction task.}
    \label{fig:bionic_overview}
\end{figure*}

Under the BIONIC framework, multimodal learning is driven by a dual latent representation that separates generative structure from task-oriented discrimination (Figure ~\ref{fig:bionic_overview}). The generative component is responsible for capturing shared and modality-specific variability across views and for explicitly modeling missing data, while the discriminative component focuses on aggregating predictive information for classification. We describe these two components in turn, starting with the generative formulation.

The generative latent variables $\mathbf{g}_n \in \mathbb{R}^H$ are assigned a standard Gaussian prior,
\begin{equation}
\mathbf{g}_n \sim \mathcal{N}(\mathbf{0}, \mathbf{I}_H),
\end{equation}
where $\mathbf{I}_{H}$ denotes the $H \times H$ identity matrix, and generate each observed view through linear mappings,
\begin{equation}
\mathbf{x}_n^{(m)} \mid \mathbf{g}_n \sim 
\mathcal{N}\left(\mathbf{g}_n \mathbf{V}^{(m)\top}, \psi_m^{-1}\mathbf{I}_{D^{(m)}}\right), \label{eq:xg}
\end{equation}
where $\mathbf{V}^{(m)} \in \mathbb{R}^{D^{(m)} \times H}$ are view-specific loadings and $\psi_m$ denotes the noise precision.

To promote interpretability and robust model selection, sparsity-inducing priors are placed on the view-specific loadings. These priors perform Automatic Relevance Determination (ARD) by adapting the effective dimensionality of the latent spaces and encouraging parsimonious representations, which is particularly important in limited-cohort settings. 

While individual dimensions of pretrained embeddings may not be directly interpretable, sparsity at the loading level enables BIONIC to disentangle shared and modality-specific latent factors and to regularize the contribution of high-dimensional representations. This structural regularization improves robustness and uncertainty calibration, while supporting principled modality and representation selection at the level of latent factors rather than raw input dimensions.

Within this formulation, missing data are handled intrinsically by the generative component rather than through external preprocessing. When variables or entire views are unobserved for a given patient, their predictive distributions are obtained by marginalizing over the variational posterior of the generative latent variables, yielding closed-form Gaussian reconstructions. This mechanism naturally supports both variable-level and view-level missingness and enables probabilistic imputation informed by all available modalities. Crucially, imputation is coupled to downstream prediction through joint inference, such that available labels and observed views jointly shape the latent representation, allowing partially observed and unlabeled samples to contribute coherently to learning.

Complementing the generative component, the task-oriented latent variables $\mathbf{z}_n \in \mathbb{R}^K$ aggregate predictive information from the observed views,
\begin{equation}
\mathbf{z}_n \sim 
\mathcal{N}\left(
\sum_{m \in \mathcal{M}_a} \mathbf{x}_n^{(m)} \mathbf{W}^{(m)\top},
\tau^{-1}\mathbf{I}_K
\right),
\end{equation}
where $\mathbf{W}^{(m)} \in \mathbb{R}^{D^{(m)} \times K}$ are discriminative loadings subject to sparsity-inducing ARD priors. These priors enable automatic selection of task-relevant modalities and latent factors, preventing diffuse contributions from high-dimensional embeddings and improving generalization in limited-cohort settings. This probabilistic formulation yields calibrated predictive uncertainty, which is critical for risk-sensitive clinical decision support.

Finally, an intermediate random variable $t_n$ acts as an additional output view jointly generated from both latent spaces that captures task-related predictive structure,
\begin{equation}
\mathbf{t}_n = 
\mathbf{z}_n \mathbf{U}^\top +
\mathbf{g}_n \mathbf{V}^{(T)^\top} +
\boldsymbol{\epsilon}_t,
\end{equation}
with the class label $\mathbf{y}_n$ obtained via a Bayesian logistic regression conditioned on $\mathbf{t}_n$. This naturally supports SS learning by allowing unlabeled samples to contribute through the generative pathway, maintaining task-specific predictions through the discriminative latent space.
Inference is performed using a mean-field variational approximation. 

\subsection{Pretrained embeddings as multimodal views}
In BIONIC, each data modality is represented as a view in the Bayesian multi-view formulation. A view may correspond either to structured clinical variables (e.g., demographics or laboratory measurements) or to a fixed-dimensional embedding obtained offline from a pretrained encoder. High-dimensional raw modalities such as medical images, histopathology slides or clinical text are therefore not modeled directly; instead, their nonlinear structure is captured by pretrained encoders, and the resulting embeddings are treated as views alongside structured inputs. Pretrained encoders are used as deterministic feature extractors and are not updated during training, ensuring that all modalities are handled uniformly within the probabilistic framework.


Differences in dimensionality across modalities are handled via view-specific loading matrices in both the generative ($\mathbf{V}^{(m)}$) and discriminative ($\mathbf{W}^{(m)}$) components. Sparsity-inducing ARD priors placed on these loadings regulate the contribution of each modality and latent factor, enabling automatic selection of task-relevant views and adaptive control of model complexity.

When operating on pretrained embeddings, whose informative variability in limited-cohort clinical settings often lies in a low-dimensional subspace, we further support this Bayesian selection through a fixed orthogonal reparameterization of the input space prior to probabilistic modeling. This rotation provides a more suitable basis for ARD priors, allowing irrelevant directions to be effectively shrunk while preserving the full expressive capacity of the embeddings. Importantly, no dimensionality is fixed a priori: representation compactness emerges naturally from the interaction between the rotated embedding space and the sparsity-inducing priors, leading to improved generalization and stability without manual tuning.


Finally, all views—whether structured variables or pretrained embeddings—integrate seamlessly into the probabilistic formulation under missing data. If a view is unavailable for a given sample, it is treated as unobserved and marginalized during inference. Partially observed structured views and available embedding views jointly inform the posterior over latent variables, allowing BIONIC to integrate heterogeneous information while explicitly accounting for uncertainty arising from missing data.

\subsection{Interpretability pipeline}

Interpretability in deep multimodal models is hindered by nonlinear, high-dimensional pretrained representations. In BIONIC, the linear and unidirectional discriminative pathway yields an intrinsic and analytically tractable interpretability mechanism, enabling propagation of discriminative relevance from the latent space to embedding representations and, when available, to the original input domain.

After a linear orthogonal reparameterization of the embedding space and component-wise normalization, the expected model output is linear in the transformed features. Accounting for this preprocessing, the sensitivity of the expected output with respect to the embedding of modality $m$ is obtained via the chain rule as
\begin{equation}
\mathbf{S}^{(m)}
= \frac{\partial \mathbb{E}[y]}{\partial \mathbf{x}^{(m)}}
= \mathbf{R}^{(m)} \mathbf{D}^{(m)-1} \mathbf{W}^{(m)} \mathbf{U},
\end{equation}
where $\mathbf{R}^{(m)}$ is the orthogonal linear transformation that rotates the embedding space, $\mathbf{D}^{(m)}$ contains the normalization standard deviations, and $\mathbf{W}^{(m)}$ and $\mathbf{U}$ are the learned discriminative weight matrices. 
The resulting vector $\mathbf{S}^{(m)} \in \mathbb{R}^{D_m}$ defines a global discriminative sensitivity direction in the embedding space of modality $m$, reflecting how variations along each embedding direction influence the model output on average.

To obtain sample-level relevance, we can project the sample embedding onto the discriminative sensitivity direction to define a relevance score that captures the aligned contribution of a sample modality to the discriminative task through the task-oriented pathway.


\section{Materials}

We evaluate the proposed framework on three publicly available multimodal clinical datasets with heterogeneous modality availability and missing data patterns. Table~\ref{tab:datasets_summary} summarizes the number of patients, modalities, dimensionality, and missing data statistics for each dataset.

\begin{table}[t]
\caption{Description of the datasets used in this study. We report the number of patients and data modalities, together with the proportion of missing data for either entire modalities or missing features within observed modalities. The dimensionality of each modality is also indicated.}
\centering
\small
\begin{tabular}{cccccccc}
\toprule
\multirow{2}{*}{\textbf{Dataset}} &
\multirow{2}{*}{\textbf{Patients}} &
\multirow{2}{*}{\textbf{Modality}} &
\multicolumn{2}{c}{\textbf{Missing (\%)}} &
\multirow{2}{*}{\textbf{Dimensions}} \\
\cmidrule(lr){4-5}
& & &
\textbf{Modality} &
\textbf{Feature} &
\\
\midrule

\multirow{5}{*}{MMIST} & 
\multirow{5}{*}{618} & 
CT & 61.3 & --- & 448 x 448 x 56 \\
& & MRI & 91.6 & --- & 448 x 448 x 56 \\
& & WSI & 62.6 & --- & 256×256 (per patch) \\
& & Genomic & 25.2 & --- & 3 \\
& & Clinical & --- & 14.1 & 14 \\
\midrule

\multirow{2}{*}{MOTUM} & 
\multirow{2}{*}{67} & 
MRI & 4.5 & --- & 260 x 320 x 21 \\
& & Clinical & --- & 11.4 & 5 \\
\midrule

\multirow{3}{*}{TCGA} & 
\multirow{3}{*}{867} & 
WSI & --- & --- & 256x256 (per patch) \\
& & Text & --- & --- & 400 \\
& & Transcrip. & --- & --- & 18981 \\
\bottomrule
\end{tabular}
\label{tab:datasets_summary}
\end{table}

\subsubsection{MMIST}

MMIST \cite{mmist_ccrcc} includes 618 patients diagnosed with clear cell renal cell carcinoma and comprises five modalities: Computed Tomography (CT), Magnetic Resonance Imaging (MRI), Whole Slide Images (WSI), somatic mutation data for key driver genes, and structured clinical data. 
The prediction task is a 12-month survival classification. A Semi-Supervised (SS) setting is additionally considered by incorporating 47 unlabeled patients contributing WSI features only.

\subsubsection{MOTUM}

The MOTUM dataset \cite{motum} contains 67 patients with either high-grade gliomas or brain metastases from six primary tumor origins. Four MRI modalities (high axial resolution) were used: Fluid-Attenuated Inversion Recovery (FLAIR), T1-weighted, contrast-enhanced T1-weighted (T1ce), T2-weighted; and clinical metadata.
The considered classification task was binary discrimination between glioma and metastasis. MRI scans were preprocessed following standard neuroimaging pipelines, and clinical variables include age, sex, extent of resection, P53 status and the Ki-67 index. All features were standardized prior to modeling.

\subsubsection{TCGA}
For the TCGA dataset, we used the Breast Cancer (BRCA) subset, processed by the authors of \cite{ren_otsurv_2025} and generated by the TCGA Research Network \cite{koboldt_comprehensive_2012}. 
This consists of histological WSI samples sourced from 867 patients with breast cancer, diagnostic question-answer pairs obtained from the WSI-VQA dataset \cite{chen_wsi-vqa_2025} as well as transcriptomic.
The prediction task is defined as a binary classification distinguishing between Basal/triple-negative (the most aggressive) and other breast cancer subtypes. A SS setting is additionally considered with the incorporation of 20 unlabeled patients.

\section{Results and Discussion}
\label{sec:results}

\subsection{Experimental Setup}

\begin{table}[t]
\caption{Summary of embeddings for each dataset.}
\label{tab:embeddings}
\centering
\small
\begin{tabular}{cccc}
\toprule
\textbf{Dataset} & \textbf{Modality} & \textbf{Embedding model} & \textbf{Dim.} \\
\midrule
\multirow{4}{*}{MMIST}
& CT & MedicalNet \cite{chen2019med3d} & 512 \\
& MRI & MedicalNet \cite{chen2019med3d} & 512 \\
& WSI & CLAM  \cite{lu2021clam} & 1024 \\
\midrule
\multirow{1}{*}{MOTUM}
& MRI & AMAES \cite{munk_amaes_2024} & 512x8x8x8 \\
\midrule
\multirow{3}{*}{TCGA}
& WSI & UNI \cite{chen_towards_2024} & 1024 (per patch) \\
& Transcriptomic & BulkRNABert \cite{pmlr-v259-gelard25a} & 256 \\
& Text & BioBART \cite{yuan2022biobart} & 400x768 \\
\bottomrule
\end{tabular}
\end{table}

High-dimensional raw modalities were converted into compact representations using domain-specific pretrained models. Table~\ref{tab:embeddings} summarizes the embedding models and dimensionality used for each dataset and modality. For MMIST and for the WSI modality of TCGA, we utilize the pretrained embeddings provided by the authors \cite{mmist_ccrcc, ren_otsurv_2025}.

To evaluate the effectiveness of our approach, we conduct a comprehensive comparison against state-of-the-art multimodal methods, including both classical Machine Learning (ML) techniques and recent Deep Learning (DL) models. Specifically, we benchmark BIONIC against Multiple Kernel Learning - Support Vector Machine (MKL-SVM) with a Gaussian kernel \cite{zhang2024explaining}, multimodal Partial Least Squares (M-PLS) \cite{belenguer2024unified}, multimodal Factor Analysis (M-FA) \cite{sevilla2022bayesian}, Late-Fusion SVM (LF-SVM) \cite{imrie2025automated}, and six deep multimodal baselines: Cross-modal Prediction Model (CPM) \cite{zhang2019cpm}, Trusted Multi-view Classification (TMC) \cite{han2022trusted}, Cross-Supervised Multimodal Variational Information Bottleneck (CSMVIB) \cite{zhang2025towards}, Deep Incomplete Multi-View (DeepIMV) \cite{lee2021variational}, Reliable Cross-Modal Learning (RCML) \cite{xu2024reliable}, and Trusted Uncertainty-aware Deep learning (TUNED) \cite{huang2025trusted}. 

Regarding the hyperparameter selection criterion, for the Bayesian models (M-PLS, M-FA), no hyperparameters were cross-validated, as they are trained via variational inference. For the DL models, we used author-recommended ranges. Furthermore, for the MKL-SVM and LF-SVM, the regularization parameter was cross-validated using an inner 10-fold stratified cross-validation from $10^{-3}$ to $10^{3}$ in log-scale steps. Moreover, missing values were handled using median imputation for all models, except for those that natively support incomplete data (M-FA, CPM, DeepIMV, and BIONIC).

M-FA and BIONIC were initialized with 100 latent dimensions in the generative space, while M-PLS and BIONIC used a discriminative latent dimensionality equal to the number of output classes minus one\footnote{As in Bayesian Partial Least Squares (BPLS), the number of discriminative latent components $K$ is bounded by $C-1$ for a $C$-class classification problem.}. For BIONIC, sparsity-inducing priors automatically pruned irrelevant dimensions from the embedding space. To promote this pruning, we rotated the embedding space along maximum-variance directions (preserving $99.9\%$ of the original variance), yielding compact, task-adaptive representations without manual tuning. All Bayesian models (BIONIC, M-PLS, and M-FA) were trained using variational inference and were considered converged when the lower bound satisfied $L(q)_{T-100} > L(q)_T(1 - 10^{-8})$.

Model performance was evaluated using non-overlapping, class-stratified 10-fold cross-validation. We report the mean and standard deviation across folds for both Area Under the ROC Curve (AUC) and Balanced Accuracy (BACC).



\subsection{Discriminative Performance}

\begin{table*}[!th]
\centering
\caption{Classification performance of BIONIC (rightmost column) compared to all baseline methods. For each dataset and model, we report the AUC (white cells) and  BACC (light gray cells).}
\setlength{\tabcolsep}{2pt}
\begin{adjustbox}{max width=\textwidth}
    \begin{tabular}{cccccccccccc}
    \toprule
    & MKL-SVM &  M-PLS & M-FA & LF-SVM & CPM & TMC & CSMVIB & DeepIMV & RCML & TUNED & BIONIC\\ \midrule
    \multirow{2}{*}{MMIST}    
                & \cellcolor{white!100} 0.73 $\pm$ 0.03 
                & \cellcolor{white!100} 0.63 $\pm$ 0.06 
                & \cellcolor{white!100} 0.60 $\pm$ 0.06 
                & \cellcolor{white!100} 0.63  $\pm$ 0.07 
                & \cellcolor{white!100} 0.70  $\pm$ 0.04 
                & \cellcolor{white!100} 0.72 $\pm$ 0.03 
                & \cellcolor{white!100} 0.65 $\pm$ 0.04 
                & \cellcolor{white!100} 0.54 $\pm$ 0.05 
                & \cellcolor{white!100} 0.73 $\pm$ 0.04 
                & \cellcolor{white!100} 0.66 $\pm$ 0.04 
                & \cellcolor{white!100} \textbf{0.84} $\pm$ \textbf{0.03} \\
                & \cellcolor{gray!10} 0.51  $\pm$ 0.03  
                & \cellcolor{gray!10}  0.52 $\pm$ 0.05 
                & \cellcolor{gray!10} 0.49 $\pm$ 0.03  
                & \cellcolor{gray!10} 0.56 $\pm$ 0.06  
                & \cellcolor{gray!10} 0.66 $\pm$ 0.05 
                & \cellcolor{gray!10} 0.68 $\pm$ 0.02  
                & \cellcolor{gray!10} 0.61 $\pm$ 0.05  
                & \cellcolor{gray!10} 0.51 $\pm$ 0.04  
                & \cellcolor{gray!10} 0.66 $\pm$ 0.03  
                & \cellcolor{gray!10} 0.65 $\pm$ 0.03  
                & \cellcolor{gray!10}  \textbf{0.79} $\pm$ \textbf{0.03}  \\

    \multirow{2}{*}{MOTUM}    
                & \cellcolor{white!100} 0.80 $\pm$ 0.17
                & \cellcolor{white!100} 0.77 $\pm$ 0.10
                & \cellcolor{white!100} 0.49 $\pm$ 0.18
                & \cellcolor{white!100} 0.85 $\pm$ 0.14
                & \cellcolor{white!100} 0.71 $\pm$ 0.18
                & \cellcolor{white!100} 0.69 $\pm$ 0.21
                & \cellcolor{white!100} 0.73 $\pm$ 0.20
                & \cellcolor{white!100} 0.51 $\pm$ 0.23
                & \cellcolor{white!100} 0.48 $\pm$ 0.19
                & \cellcolor{white!100} 0.55 $\pm$ 0.28
                & \cellcolor{white!100} \textbf{0.87} $\pm$ \textbf{0.11} \\
                & \cellcolor{gray!10} 0.75 $\pm$ 0.13  
                & \cellcolor{gray!10}  0.70 $\pm$ 0.13   
                & \cellcolor{gray!10} 0.5 $\pm$ 0.0   
                & \cellcolor{gray!10}  0.79 $\pm$ 0.10   
                & \cellcolor{gray!10}  0.67 $\pm$ 0.19   
                & \cellcolor{gray!10} 0.63 $\pm$ 0.15   
                & \cellcolor{gray!10}  0.65 $\pm$ 0.21  
                & \cellcolor{gray!10}  0.5 $\pm$ 0.0  
                & \cellcolor{gray!10}   0.5 $\pm$ 0.0   
                & \cellcolor{gray!10}   0.5 $\pm$ 0.0   
                & \cellcolor{gray!10} \textbf{0.83} $\pm$ \textbf{0.09} \\
                
    \multirow{2}{*}{TCGA}    
                & \cellcolor{white!100} 0.92 $\pm$ 0.03 
                & \cellcolor{white!100} 0.94 $\pm$ 0.03 
                & \cellcolor{white!100} 0.97 $\pm$ 0.01 
                & \cellcolor{white!100} 0.93 $\pm$ 0.02 
                & \cellcolor{white!100} \textbf{0.98} $\pm$ \textbf{0.02} 
                & \cellcolor{white!100} \textbf{0.98} $\pm$ \textbf{0.02} 
                & \cellcolor{white!100} \textbf{0.98} $\pm$ \textbf{0.02} 
                & \cellcolor{white!100} \textbf{0.98} $\pm$ \textbf{0.01} 
                & \cellcolor{white!100} \textbf{0.98} $\pm$ \textbf{0.02} 
                & \cellcolor{white!100} \textbf{0.98} $\pm$ \textbf{0.01} 
                & \cellcolor{white!100} \textbf{0.98} $\pm$ \textbf{0.01}\\
                & \cellcolor{gray!10} 0.5 $\pm$ 0.0 
                & \cellcolor{gray!10} 0.82 $\pm$ 0.04  
                & \cellcolor{gray!10} 0.91 $\pm$ 0.03  
                & \cellcolor{gray!10} 0.88 $\pm$ 0.02  
                & \cellcolor{gray!10} 0.92 $\pm$ 0.02  
                & \cellcolor{gray!10} 0.93 $\pm$ 0.03  
                & \cellcolor{gray!10}  0.90 $\pm$ 0.03  
                & \cellcolor{gray!10} 0.90 $\pm$ 0.03  
                & \cellcolor{gray!10} 0.91 $\pm$ 0.02  
                & \cellcolor{gray!10} 0.92 $\pm$ 0.03  
                & \cellcolor{gray!10} \textbf{0.96} $\pm$ \textbf{0.02} \\
    \bottomrule
    \end{tabular}
    \end{adjustbox}
    \label{tab:results_standard}
\end{table*}

Table \ref{tab:results_standard} reports the discriminative performance of BIONIC compared to a broad set of classical, shallow, and deep multimodal baselines across three datasets, using AUC and BACC as evaluation metrics. Overall, BIONIC consistently achieves the best or among the best performances across datasets, demonstrating robust discriminative capabilities.

On MMIST, BIONIC substantially outperforms all baselines, improving both AUC and BACC by $0.11$ over the strongest competitors; this 
corresponds to an improvement of over 15\% relative to the strongest baseline, which is particularly significant given the severe modality-level missingness in MMIST.
While only M-FA, CPM, and DeepIMV natively support incomplete data, BIONIC demonstrates superior robustness, yielding more balanced predictions and a clearer decision boundary, as reflected by its higher BACC. These results highlight the benefit of explicitly modeling missingness within a unified generative–discriminative framework. On MOTUM, our proposal again achieves the best results, outperforming LF-SVM, which is the strongest baseline in this low-sample regime. Consistent with prior observations, linear methods tend to outperform deep models due to reduced overfitting; yet, BIONIC shows the largest relative gain on this dataset, underscoring the regularising effect of its latent generative structure and Bayesian formulation. On TCGA, where all methods achieve high AUC due to the relative ease of the task, BIONIC matches the best AUC while yielding the highest BACC. In contrast to several baselines with high AUC but lower BACC, BIONIC maintains balanced accuracy, indicating improved calibration and robustness across classes.

\subsection{Generative Performance}

To assess BIONIC’s robustness to unlabeled samples enabled by its generative formulation, we evaluate its performance under three supervision conditions. In the Supervised (S) setting, the model is trained exclusively using labeled training samples. In the SS setting, additional unlabeled samples from the training data are incorporated during training. Finally, in the Transductive SS (TSS) setting, the model is trained using all available input samples, including test inputs, while strictly excluding test labels. This setting reflects a realistic deployment scenario in which newly acquired, unlabeled data can be incorporated via fast inference to refine the latent space before producing predictions.


\begin{table}[!th]
\centering
\caption{Classification performance for each dataset of BIONIC under 3 supervision settings: Supervised (S), Semi-Supervised (SS), and Transductive Semi-Supervised (TSS). We report the AUC (white cells) and  BACC (light gray cells)}
    \begin{tabular}{c c c c}
    \toprule
    \multirow{2}{*}{~}  & \multicolumn{3}{c}{Supervision type} \\
    & S & SS & TSS  
    \\ \midrule
    
    \multirow{2}{*}{MMIST}    
                & \cellcolor{white!100} 0.81 $\pm$ 0.02  
                & \cellcolor{white!100} 0.84 $\pm$ 0.03 
                & \cellcolor{white!100} \textbf{0.90} $\pm$  \textbf{0.02} \\
                & \cellcolor{gray!10} 0.75 $\pm$ 0.02
                & \cellcolor{gray!10} 0.79 $\pm$ 0.03
                & \cellcolor{gray!10} \textbf{0.85} $\pm$ \textbf{0.02}  \\

    \multirow{2}{*}{MOTUM}   
                & \cellcolor{white!100} 0.87 $\pm$ 0.11 
                & \cellcolor{white!100} --- 
                & \cellcolor{white!100} \textbf{0.92} $\pm$ \textbf{0.15}   \\
                & \cellcolor{gray!10} 0.83  $\pm$ 0.09
                & \cellcolor{gray!10}  ---
                & \cellcolor{gray!10} \textbf{0.90} $\pm$ \textbf{0.18}   \\
                
    \multirow{2}{*}{TCGA}    
                & \cellcolor{white!100} \textbf{0.98} $\pm$ \textbf{0.01} 
                & \cellcolor{white!100} \textbf{0.98} $\pm$ \textbf{0.01} 
                & \cellcolor{white!100} \textbf{0.98} $\pm$ \textbf{0.01}  \\
                & \cellcolor{gray!10} 0.94 $\pm$ 0.02
                & \cellcolor{gray!10} 0.96 $\pm$ 0.02
                & \cellcolor{gray!10} \textbf{0.98} $\pm$ \textbf{0.02}  \\
                
    \bottomrule
    \end{tabular}
    \label{tab:results_semi}
\end{table}

Table~\ref{tab:results_semi} summarizes the test-set performance of BIONIC under the three supervision settings. Overall, incorporating unlabeled samples during training consistently improves performance over the fully supervised baseline, demonstrating the benefit of the generative component in exploiting additional data without requiring label information.

On MMIST, performance improves monotonically as supervision constraints are progressively relaxed. The transition from S to SS training yields a clear gain, while the TSS setting leads to a substantial improvement of approximately $0.1$ in AUC with respect to the supervised case. This result highlights the ability of the generative model to effectively benefit from additional unlabeled inputs to refine the latent space and improve downstream predictions. For MOTUM, where no additional unlabeled training samples are available, performance in the S and TSS settings remains comparable, with a consistent improvement observed in the transductive case. Despite the higher variability due to the limited sample size, the observed gains indicate that even modest increases in available unlabeled data can positively impact performance when integrated through the generative framework. In the case of TCGA, where overall performance is already near saturation, differences in AUC across supervision settings are minimal. However, consistent improvements are observed in balanced accuracy when unlabeled samples are incorporated, particularly under transductive training. This suggests that the generative component contributes to more balanced and better-calibrated predictions, even when discriminative performance is already high.

\subsection{Interpretability Analysis}
\label{subsec:interpretability}


In this section, we analyze the interpretability capability of BIONIC. We illustrate the sensitivity-based analysis on two modalities: MRI embeddings from the MOTUM database and text embeddings from the TCGA database. Besides, an analysis of the learnt weights is available in the Supplementary Material.

Figure \ref{fig:sens_mri} illustrates the anatomical patterns associated with the sensitivity direction of the MRI embeddings. The relevance maps were generated by decoding the embedding of a representative patient perturbed along this direction, and subtracting the baseline reconstruction. The resulting intensity difference maps are overlaid onto the subject's reconstructed anatomy to highlight regions of increased (red) or decreased (blue) signal intensity.

The sparsity-inducing structure of the model identified T1ce and FLAIR as the most discriminative modalities. This selection is clinically plausible in the context of brain metastasis, as both sequences are sensitive to vascular and perfusion-related alterations: T1ce highlights blood–brain barrier disruption through contrast enhancement, while FLAIR suppresses cerebrospinal fluid signal and emphasizes pathological tissue associated with abnormal perfusion.

Figure ~\ref{fig:sens_mri} shows that, rather than concentrating on peritumoral or surrounding regions, the relevance maps exhibit a more scattered distribution across the brain.
This behavior is expected as the morphology of both primary and secondary tumors is often indistinguishable through MRI \cite{gong2024multi}. Thus, the model relied mainly in extra-tumoral more general brain regions.

Negative mask (blue) values associated with metastasis formed consistent clusters in both FLAIR and T1ce. Specifically in Figure ~\ref{fig:sens_mri}(a,c), we can observe them through the cerebellum and brainstem, while in Figure ~\ref{fig:sens_mri}(b,d) across the corpus callosum and the transition between the cortex gray matter and the subcortical white matter (cortico-subcortical junction). These regions correspond to areas of high or terminal vascularization \cite{schmahmann2008cerebral, mohammadi2024brain}, consistent with the hematogenous spread and expansile growth pattern of metastatic cells, which lack the ability to infiltrate cortical parenchyma. In contrast, positive masks (red) associated with primary tumors were predominantly located in the cerebral cortex and peri/interventricular regions in both T1ce and FLAIR (Figure ~\ref{fig:sens_mri}(b,d)). This distribution reflects the infiltrative behavior of primary glial tumors, which arise from native brain cells and spread through white matter tracts and subependymal surfaces to ventricular and cortical regions \cite{weller2024glioma, van2023primary}. 


\begin{figure}[!th]
    \centering
    \begin{subfigure}{0.4\columnwidth}
        \includegraphics[width=\columnwidth]{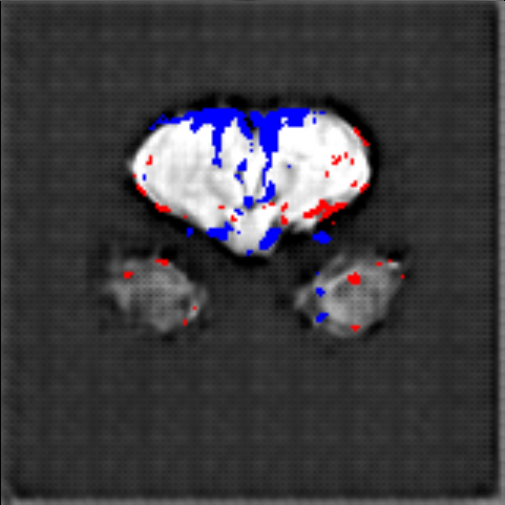}
        \caption{FLAIR (z = 32)}
    \end{subfigure}
    \begin{subfigure}{0.4\columnwidth}
        \includegraphics[width=\columnwidth]{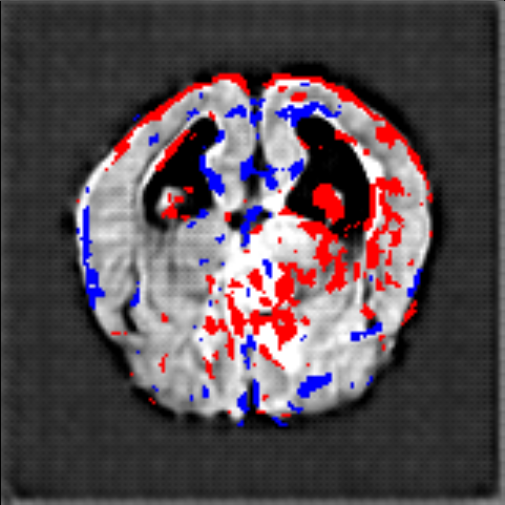}
        \caption{FLAIR (z = 81)}
    \end{subfigure}

    \vspace{1em}

    \begin{subfigure}{0.4\columnwidth}
        \includegraphics[width=\columnwidth]{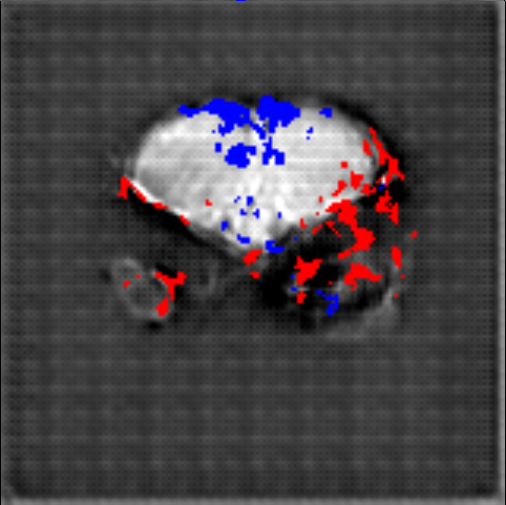}
        \caption{T1ce (z = 32)}
    \end{subfigure}
    \begin{subfigure}{0.4\columnwidth}
        \includegraphics[width=\columnwidth]{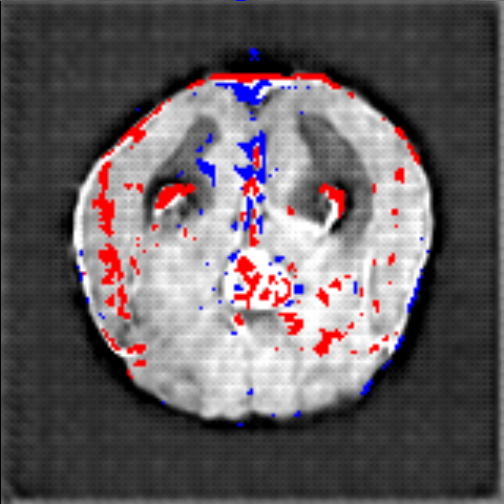}
        \caption{T1ce (z = 81)}
    \end{subfigure}
    \caption{Voxel-wise relevance map obtained via sensitivity analysis, overlaid onto a representative reconstructed patient scan (axial view). Positive intensity changes (red) are associated with primary tumor and negative values (blue) are associated with metastasis.
    }
    
    \label{fig:sens_mri}
\end{figure}



\begin{figure}[t!]
    \centering
    \begin{subfigure}[t]{0.7\textwidth}
        \centering
        \includegraphics[width=\columnwidth]{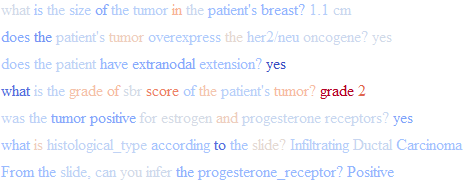}
        \caption{Non basal patient}
        \label{fig:nonBasal}
    \end{subfigure}%
    \\
    \begin{subfigure}[t]{0.7\textwidth}
        \centering
        \includegraphics[width=\columnwidth]{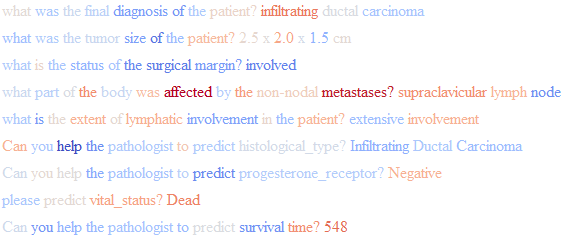}
        \caption{Basal patient}
        \label{fig:Basal}
    \end{subfigure}
    \caption{Token-level discriminative relevance for representative non-basal (top) and basal (bottom) patients. 
Red denotes positive relevance and blue negative relevance, with color intensity proportional to relevance magnitude.}
    \label{fig:sens_text}
\end{figure}

For the text modality, token-level embeddings were used to compute relevance scores for two relevant patients, Figure ~\ref{fig:sens_text}. The non-basal case (Figure ~\ref{fig:nonBasal}) shows high relevance on tokens describing tumor grade and size (e.g., \texttt{grade}, \texttt{score}, numeric dimensions), while function words receive low or mixed relevance. The basal case (Figure ~\ref{fig:Basal}) highlights tokens related to tumor extent, systemic involvement, and aggressiveness (e.g., \texttt{affected}, \texttt{metastases}, \texttt{supraclavicular}, \texttt{lymphatic involvement}, survival time), with routine question tokens down-weighted.

Tokens for the question ``\textit{can you infer the progesterone receptor?}'' reflect the classification task: positive answers in non-basal patients yield negative relevance toward basal, whereas negative answers in basal patients yield positive relevance, indicating that the model captures anti-correlated predictive relationships. Overall, relevance concentrates on clinically informative content tokens rather than generic syntax. However, note that token-level attributions reflect discriminative behavior of the model and should be interpreted qualitatively, as they depend on contextual interactions within the transformer.

\section{Conclusions}
\label{sec:conclussions}

In this work, we introduced BIONIC, a Bayesian framework for integrating heterogeneous multimodal clinical data under structured missingness. By combining a generative–discriminative latent architecture with sparsity-inducing priors, BIONIC addresses key challenges in real-world clinical settings, including multimodal data integration, modality- and variable-level missingness, limited cohort sizes, and the need for interpretable predictions. The central strength of the framework is its ability to integrate structured clinical variables with complex unstructured modalities via pretrained embeddings, without retraining high-capacity encoders. Missing data are handled natively through multimodal latent inference, enabling principled imputation by utilizing information across available modalities and, when present, label-related signals. This joint modeling strategy ensures robustness in partially observed and semi-supervised scenarios that closely reflect realistic clinical acquisition processes.

Experimental results on three multimodal clinical datasets demonstrate that BIONIC achieves strong and consistent discriminative performance across varying data availability and supervision regimes, with particular advantages under modality-level missingness and in low-sample settings where purely discriminative approaches tend to degrade. Beyond predictive performance, the framework provides intrinsic interpretability mechanisms that enable attribution of modality, embedding and feature relevance, supporting downstream clinical analysis.

A practical limitation of the proposed approach is its dependence on the quality of pretrained embeddings, as the clinical relevance of the inferred representations is inherently tied to the capacity of the underlying foundation models. When modality-specific decoders are available, sensitivity measures can be propagated back to the original input domains, enabling more intuitive interpretations in clinically meaningful feature spaces.
Future work will explore extensions including adaptive or uncertainty-aware embedding mechanisms, explicit modeling of temporal structure for longitudinal data, and further investigation of interpretability in clinical decision-support settings, as well as patient-specific explanations to facilitate real-world deployment.

\bibliographystyle{IEEEtran}
\bibliography{bibliography}
\end{document}